\documentclass{article} 
\usepackage[T1]{fontenc}
\usepackage[table]{xcolor}
\usepackage{iclr2027_conference,times}
\usepackage{amsmath}
\usepackage{booktabs}
\usepackage{tabularx}
\usepackage{capt-of}

\usepackage{amsmath,amsfonts,bm}

\def\eqref#1{equation~\ref{#1}}

\def\1{\bm{1}}

\DeclareMathAlphabet{\mathsfit}{\encodingdefault}{\sfdefault}{m}{sl}
\SetMathAlphabet{\mathsfit}{bold}{\encodingdefault}{\sfdefault}{bx}{n}

\usepackage{amssymb}
\usepackage{graphicx}
\usepackage{microtype}
\usepackage{xurl}
\usepackage{multirow}
\usepackage{adjustbox}
\usepackage[hidelinks]{hyperref}

\newcommand{\tableresize}{0.95}

\AddToHook{env/table/begin}{%
  \setlength{\abovecaptionskip}{0pt}%
  \setlength{\belowcaptionskip}{5pt}%
}

\title{\raggedright PanoVLN: Towards Effective Panoramic Vision-and-Language Navigation}

\author{
\normalfont
Zhen Wang$^{1}$ \quad
Changpeng Wang$^{1}$ \quad
Zhe Liu$^{2}$ \quad
Zhangyang Qi$^{2}$ \\
Yuxiang Lu$^{2}$ \quad
Zimo Zeng$^{1}$ \quad
Donglian Qi$^{1}$ \quad
Xi Chen$^{2}$ \\
\\[-0.5em]
$^{1}$Zhejiang University
\qquad
$^{2}$The University of Hong Kong
}

\iclrfinalcopy 
\providecommand{\panovlnfigure}[2]{%
  \IfFileExists{#1}{%
    \includegraphics[width=\linewidth,height=#2,keepaspectratio]{#1}%
  }{%
    \fbox{\parbox[c][#2][c]{\dimexpr\linewidth-2\fboxsep-2\fboxrule\relax}{%
      \centering\small Figure placeholder}}%
  }%
}

\begin{document}
\raggedbottom 

\maketitle
\lhead{} 
\renewcommand{\headrulewidth}{0pt}

\noindent
\begin{minipage}{\linewidth}
    \centering
    \includegraphics[width=1.0\linewidth]{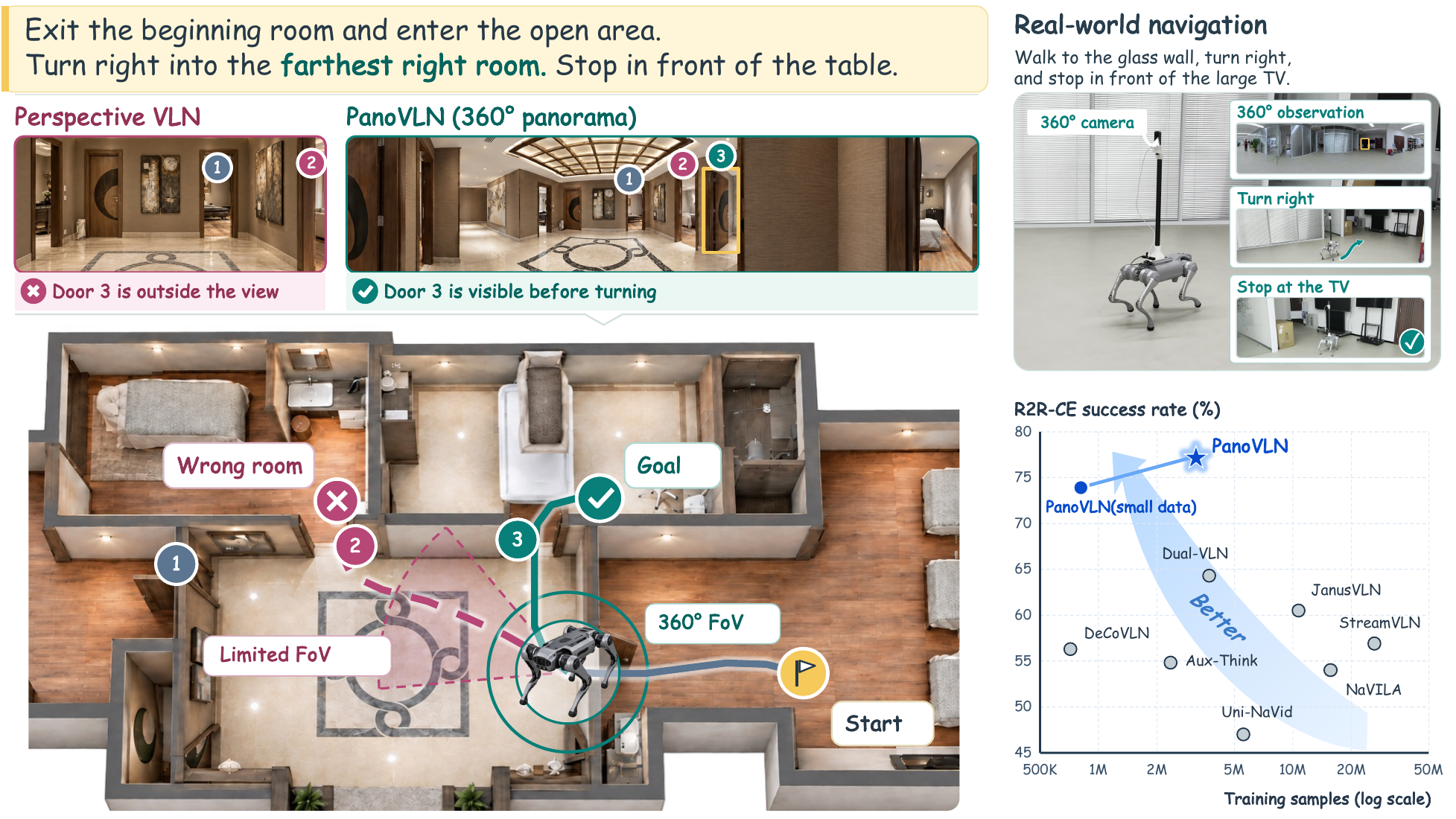}
    \captionof{figure}{PanoVLN fully exploits the wider visual context of panoramas to make vision-and-language navigation more accurate and efficient.}
    \label{fig:teaser}
\end{minipage}\par\medskip

\begingroup
\microtypesetup{activate=false}
\begin{abstract}

Recent vision-language models~(VLMs) have advanced vision-and-language navigation~(VLN), enabling models to predict navigation actions from visual observations and language instructions.
In this work, we explore VLN with panoramic observations and introduce \textbf{PanoVLN}.
The motivation is straightforward: more complete visual context should enable better-informed navigation decisions.
For example, a panorama can reveal a passage outside a perspective camera's field of view, allowing the model to identify the intended route without additional exploration.
However, we find that simply replacing perspective images with panoramas yields only limited gains.
Our diagnosis suggests that fully exploiting wider visibility requires modifications to action prediction, training supervision, and visual representation.
First, wider visibility supports longer-horizon action planning.
We make the model predict longer action sequences, enabling larger turns and subsequent movement from a single panorama.
Specifically, we introduce a confidence-guided execution~(CGE) strategy that dynamically determines how many predicted actions to execute before replanning.
Second, wider visibility also brings more complex route choices.
We therefore construct training routes with frequent branching points and clear instructions to provide targeted supervision for route selection.
Third, panoramic navigation requires understanding spatial relationships across viewing directions, beyond recognizing individual landmarks.
We combine semantic and geometric features from RGB panoramas to capture both scene content and spatial layout without adding visual tokens.
With a 4B backbone and RGB-only input, PanoVLN surpasses the previous SOTA by 11.9\% and 8.7\% in success rate on R2R-CE and RxR-CE Val-Unseen.
Real-world experiments on a quadruped further demonstrate faster navigation with fewer pauses than prior VLN methods.
Our project page is available at \url{https://wangzhen-w.github.io/PanoVLN/}.

\end{abstract}

\endgroup

\section{INTRODUCTION}
\label{sec:introduction}

Vision-and-language navigation~(VLN) requires an agent to navigate through an environment by following a natural-language instruction~\citep{r2r,r2rce}.
Recent vision-language models~(VLMs) have advanced this task by predicting navigation actions from visual observations and instructions~\citep{navid,navila,streamvln}.

Most of the previous works take perspective images as input; in this work, we explore whether panoramic observations can further improve navigation by making more of the surrounding environment available for each decision.
The motivation is straightforward: an equirectangular panorama~(ERP) provides a $360^\circ$ view, exposing passages, landmarks, and route alternatives across different viewing directions and providing a more complete visual basis for navigation decisions.

However, we find that simply replacing perspective images with ERPs under the same training and inference setup does not improve performance. 
Motivated by this observation, we conduct a detailed diagnostic analysis and find that exploiting panoramic context requires targeted adaptations. Specifically, we adapt action prediction, training supervision, and visual representation to panoramic inputs, and propose PanoVLN.
%

%

%
%

First, as a route changes direction, it may extend beyond the left or right edge of a perspective image. 
A panorama\textquoteright{}s $360^\circ$ view can show where the route continues, providing visual context for a longer sequence of navigation actions.
We therefore train PanoVLN with a longer prediction horizon, and our horizon study shows that panoramic policies favor much longer action sequences than perspective policies.
Predictions farther into the future are naturally less reliable, so executing the entire sequence is not always desirable.
We therefore introduce \emph{confidence-guided execution}~(CGE), which uses action uncertainty to determine how many predicted actions to execute before reobserving and replanning.

%
Second, a panorama is particularly useful at a branching point, where its wider field of view can reveal several possible paths.
However, branching points are relatively sparse in existing VLN training data, providing limited supervision for learning to use this advantage.
We therefore construct a dataset of 98K trajectories across 800 HM3D scenes~\citep{hm3d}, with routes that contain frequent branching points and instructions that clearly specify which path to take.
We generate and verify the instructions against route observations to ensure that the described movements and choices are visually grounded.
We further increase the sampling frequency around turns and stopping points, providing stronger supervision where route decisions and completion matter most.
Together, these choices provide dense, targeted supervision for learning route decisions and following the selected path to completion.

Third, using panoramic observations for navigation requires understanding the spatial relationships among different parts of the scene.
The VLM's visual features mainly capture scene semantics, while navigation also depends on how landmarks, passages, and other scene elements are spatially arranged.
We therefore combine the VLM's semantic features with geometric features extracted by PanoVGGT~\citep{panovggt} from the same RGB panorama.
We align and fuse features from corresponding ERP regions, providing both semantic and geometric information without additional visual tokens or depth input.

With a 4B backbone and RGB-only observations, PanoVLN achieves success rates of \(77.3\%\) on R2R-CE Val-Unseen and \(78.0\%\) on RxR-CE Val-Unseen, exceeding the previous state of the art by \(11.9\%\) and \(8.7\%\), respectively.
On a quadruped robot, PanoVLN navigates indoor and outdoor routes in less time, with fewer policy calls and pauses than prior VLN baselines.

\section{RELATED WORK}
\label{sec:related_work}

\paragraph{Vision-and-language navigation.}
Vision-and-language navigation in continuous environments (VLN-CE) extends instruction-following routes from navigation graphs to executable motion~\citep{r2r,rxr,r2rce}.
Waypoint and map-based methods predict reachable locations, maintain spatial representations, or look ahead along candidate routes to support planning and control~\citep{bridgevln,cm2,gridmm,etpnav,hnr}.
VLM-based policies use visual histories or streaming video to predict navigation commands, sometimes passing intermediate decisions to a separate execution policy~\citep{navid,uninavid,streamvln,navila,dualvln}.
Complementary work improves waypoint supervision, the scale of navigation data, and instruction generation~\citep{law,scalevln,instrugen}.
PanoVLN builds on these advances to study how full-surround visual observations support vision-and-language navigation.

\paragraph{Panoramic geometry.}
Panoramic geometry methods account for spherical projection when estimating depth and 3D scene structure.
Depth estimation has used ERP--cubemap fusion, camera-independent spherical representations, and models trained for panoramic inputs~\citep{unifuse,unik3d,da2,dap}.
Feed-forward reconstruction models jointly estimate camera and scene geometry from images, with PanoVGGT extending this approach to panoramas~\citep{vggt,panovggt}.
Navigation methods likewise represent spatial layout through cross-modal maps, grid memories, lookahead scene features, or separate spatial and semantic memories~\citep{cm2,gridmm,hnr,janusvln}.
PanoVLN brings panoramic geometry into VLN to strengthen spatial reasoning.

\section{METHOD}
\label{sec:method}

In this work, we study how to fully exploit the complete visual context provided by panoramas in VLN task.
Starting from a baseline model with panoramas as input, we make three key adaptations: 
First, to fully utilize the wider visibility, we train and execute in longer action sequences.
Second, we construct a decision-centric training dataset more suitable for panoramic settings. 
Third, we develop geometric-aware visual representations to better understand the panoramic observations.
%
%
%

\subsection{Preliminaries}
\label{sec:preliminaries}

\paragraph{Perspective baseline.}
We begin with a VLM-based policy for VLN-CE~\citep{r2rce}.
At step \(t\), the policy receives a natural-language instruction \(x\), the current perspective RGB image \(I_t\), and sampled visual history \(\mathcal{I}_{<t}\).
A visual encoder extracts image features, which are projected into the language model's embedding space and combined with instruction tokens.
The language model autoregressively predicts a sequence of actions from \(\mathcal{A}=\{\mathtt{forward},\mathtt{left},\mathtt{right},\mathtt{stop}\}\).
The policy learns from expert action sequences and, at inference, executes a fixed-length prefix of its prediction before observing again.
Movement actions use fixed translation and rotation increments.
The \(\mathtt{stop}\) action ends the episode, and success requires stopping within the benchmark's goal region.

\paragraph{Panoramic baseline.}
We obtain a panoramic baseline by replacing both current and historical perspective images with RGB equirectangular panoramas~(ERPs) during training and inference.
An ERP linearly maps a \(360^\circ\times180^\circ\) field of view to a rectangle.
The agent's heading is centered, and the left and right image boundaries are adjacent across a seam behind the agent.
The policy input becomes \(\mathcal{O}_t=(x,\mathcal{P}_{<t},P_t)\), where \(P_t\) is the current ERP and \(\mathcal{P}_{<t}\) is the sampled panoramic history.
We retain the same VLM, training trajectories, action supervision, and execution procedure.
In our experiments, this input-only replacement yields limited gains and can even reduce navigation performance, motivating the adaptations described below.

\subsection{Longer Action-Sequence Supervision and Execution}
\label{sec:action_sequence}

A panorama can show where a route continues after it turns beyond the field of view of a perspective
image. 
Short action targets use only part of this visual context for supervision. 
We therefore use longer action-sequence supervision and adapt the execution length according to prediction uncertainty.

\paragraph{Action-sequence supervision.}
At training state \(t\), the target \(\mathbf{A}_t^*=(a_{t,1}^*,\ldots,a_{t,H}^*)\) contains the next \(H\) expert actions, padded with \(\mathtt{stop}\) beyond the trajectory end.
The policy predicts these actions autoregressively.
Using teacher forcing, we minimize
\begin{equation}
\mathcal{L}_{\mathrm{act}}
=
-\frac{1}{H}
\sum_{i=1}^{H}
\log p_\theta
\left(a_{t,i}^{*}\mid\mathcal{O}_t,\mathbf{A}_{t,<i}^{*}\right),
\label{eq:action_loss}
\end{equation}
\nopagebreak[4]
where \(\mathbf{A}_{t,<i}^{*}\) contains the preceding expert actions and \(p_\theta\) is the VLM's next-token distribution.

\begin{figure}[t]
    \centering
    \includegraphics[width=\linewidth]{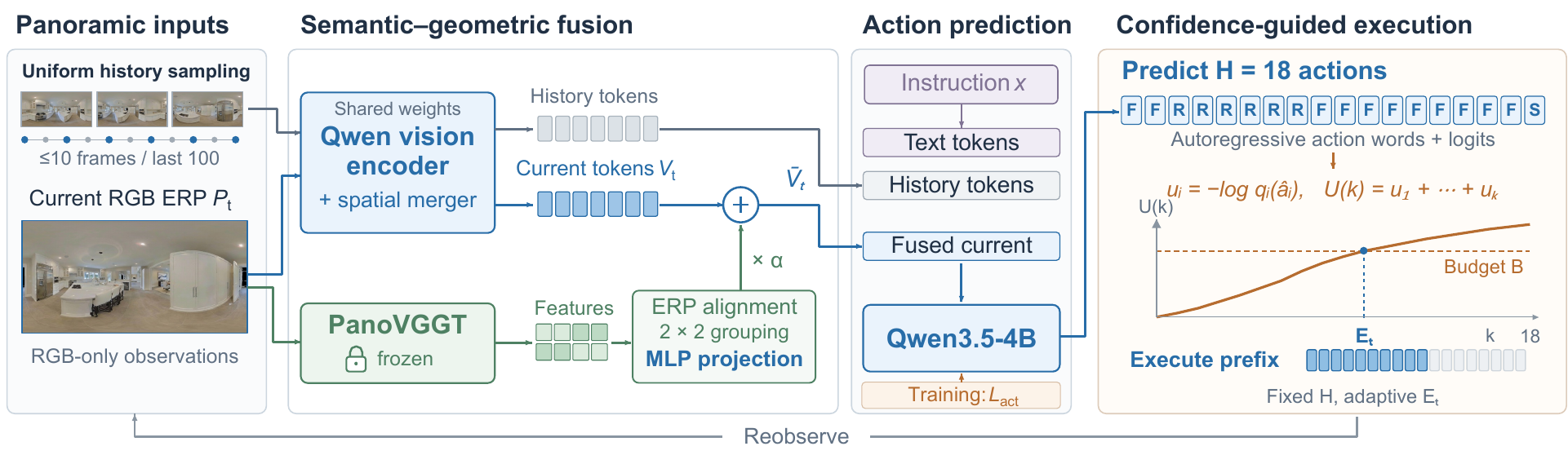}
    \caption{\textbf{PanoVLN pipeline.}
        Current and historical panoramas share a fixed visual-token budget.
        Aligned geometric features enrich the current visual tokens, and uncertainty determines how much of the predicted action sequence to execute.}
    \label{fig:policy_architecture}
\end{figure}

\noindent\begin{minipage}{\linewidth}
\begin{minipage}[t]{0.55\linewidth}
\vspace{0pt}
\textbf{Confidence-guided execution~(CGE).}
The prediction horizon \(H\) determines how far ahead the policy predicts, whereas the execution length determines when it reobserves.
CGE selects this length according to the uncertainty of the predicted actions.

Given \(\mathcal{O}_t\) and preceding predictions, let \(z_{t,i}(a)\) be the logit for action \(a\) at position \(i\).
Normalizing over \(\mathcal{A}\), we define uncertainty for the generated action \(\hat a_{t,i}\) as
\begin{equation}
\begin{aligned}
q_{t,i}(a)
&=
\frac{\exp z_{t,i}(a)}
{\sum_{b\in\mathcal{A}}\exp z_{t,i}(b)},\\
u_{t,i}
&=
-\log q_{t,i}(\hat a_{t,i}).
\end{aligned}
\label{eq:action_uncertainty}
\end{equation}
\end{minipage}\hfill
\begin{minipage}[t]{0.43\linewidth}
    \vspace{0pt}
    \centering
    \includegraphics[width=\linewidth]{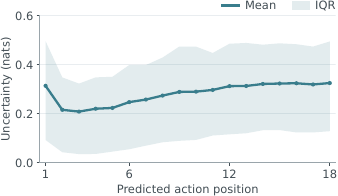}
    {\small\captionof{figure}{\textbf{Action uncertainty.} Uncertainty rises after an initial dip and varies across policy calls, motivating adaptive execution.}
    \label{fig:cge_uncertainty}}
\end{minipage}
\end{minipage}\par
Mean uncertainty rises after an initial dip, with substantial variation across policy calls (Figure~\ref{fig:cge_uncertainty}).

Let \(U_t(k)=\sum_{i=1}^{k}u_{t,i}\), with \(U_t(0)=0\).
CGE extends the prefix while \(U_t(k)\leq B\) for an uncertainty budget \(B\), selecting at least \(E_{\min}\) actions:
\begin{equation}
E_t
=
\max\left\{
k\in\{1,\ldots,H\}:
k\leq E_{\min}\ \text{or}\ U_t(k)\leq B
\right\}.
\label{eq:cge_horizon}
\end{equation}
\nopagebreak[4]
The agent executes this prefix, then reobserves unless it stops.

\subsection{Decision-Centric Data Construction}
\label{sec:training_data}

Existing VLN training data contains relatively few trajectories with frequent route choices among multiple visible paths.
This provides limited supervision for learning to select the intended path from panoramic observations.
We therefore construct trajectories with frequent branching points and pair them with instructions that clearly identify the chosen path.
We further sample turns and stopping points more densely during data construction.

\begin{figure}[t]
    \centering
    \includegraphics[width=0.95\linewidth]{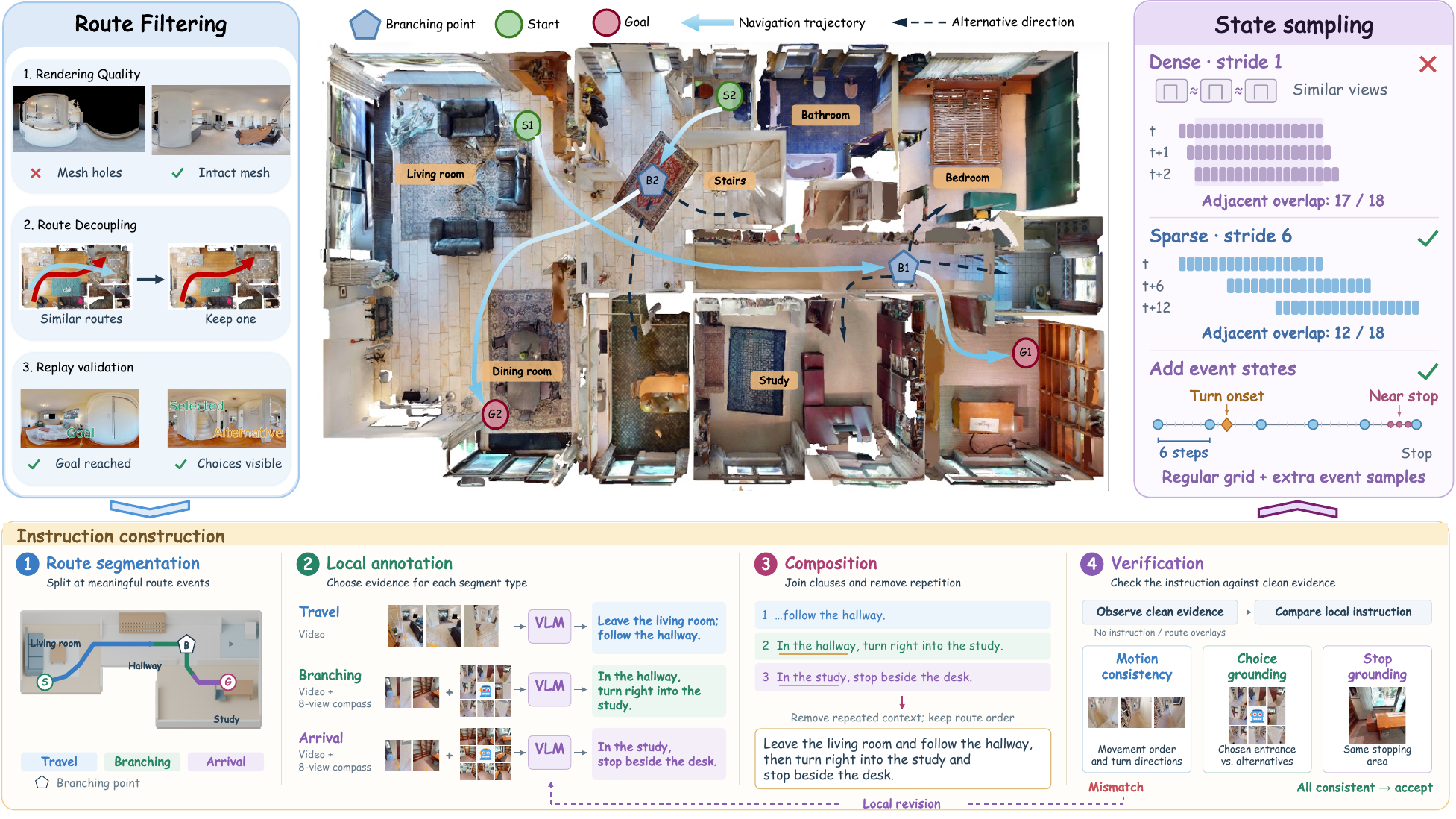}
    \caption{\textbf{Data construction pipeline}.
        We construct routes with frequent branching points, generate and verify instructions, and sample training states.}
    \label{fig:data_pipeline}
\end{figure}

\paragraph{Route construction and filtering.}
We construct 98K navigation trajectories across 800 HM3D scenes, dividing walkable space into connected areas using the navigation mesh.
A \emph{branching point} has at least two visible, traversable paths to different areas, excluding the incoming path.
We sample endpoints in different areas and retain routes through branching points.
Rendering-quality checks remove candidates with mesh holes or incomplete geometry, followed by near-duplicate removal.
An expert converts the remaining routes into primitive action sequences; replay verifies goal reachability and visibility of the chosen path and its alternatives at each branching point.

\paragraph{Instruction construction.}
We divide trajectories by route events into \emph{travel}, \emph{branching}, and \emph{arrival} segments.
Each segment uses first-person video with the expert path marked on the ground; branching and arrival also use eight-view compass images.
Qwen3.8-27B~\citep{qwen38} describes movement, identifies the chosen path from visible cues, and specifies the stopping location.
We combine descriptions in route order, remove repetition, and refine the wording.

We verify each instruction segment using clean videos and compass images without instruction or route overlays.
Motion consistency checks movement order and turn directions against the replay.
Choice grounding checks whether the instruction identifies the demonstrated entrance among alternatives; stop grounding checks whether it describes the observed arrival area.
Mismatched segments are revised locally and reverified; only samples passing all three checks are retained.

\paragraph{Training sample selection.}
Adjacent states often have similar observations and overlapping action targets.
For \(H=18\), we use a stride-six grid, reducing overlap between adjacent grid targets from 17 to 12 actions.
We add states at sustained-turn onsets and near termination to supervise turning and stopping.
Each state is paired with its \(H\)-step expert action sequence.
The grid preserves route coverage; added states emphasize action transitions.

\subsection{Geometry-aware visual representation}
\label{sec:panoramic_policy}

A panoramic observation brings different parts of the surrounding scene into a single view, making their spatial relationships important for navigation.
We therefore fuse semantic and aligned geometric features from the current RGB panorama without adding visual tokens.

\paragraph{Visual context allocation.}
We allocate \(N_c\) tokens to the current ERP and \(N_h<N_c\) tokens to each history frame.
The finer current features support route selection, while coarser historical features provide context for instruction progress.
We uniformly sample up to \(M\) past observations from a recent temporal window, giving a total visual-token budget of \(N_c+MN_h\).

\paragraph{Spatially aligned geometric fusion.}
A pretrained PanoVGGT encoder~\citep{panovggt} extracts geometric features from the current RGB panorama.
We resample them in ERP coordinates and group them to cover the same regions as the VLM's merged current tokens \(V_t\).
A trainable MLP \(f_\psi\) projects the aligned geometric groups \(G_t\) into the visual-token embedding space for residual fusion:
\begin{equation}
\bar V_t
=
V_t+\alpha f_\psi(G_t),
\label{eq:geometry_fusion}
\end{equation}
\nopagebreak[4]
where \(\alpha\) is a fixed residual scale.
Fusion combines semantics and geometry from corresponding ERP regions, preserving token count and order.
The instruction, history, and fused tokens condition action prediction.
We train the VLM and projection with Eq.~\ref{eq:action_loss} and freeze the geometry encoder.

\begin{table}[t]
    \centering
    \footnotesize
    \renewcommand{\arraystretch}{1.0}
    \setlength{\tabcolsep}{2pt}
    \caption{\textbf{Simulation benchmark comparison.}
We evaluate performance on R2R-CE and RxR-CE Val-Unseen with the indicated observation modalities.
$^*$ denotes the waypoint predictor from~\citet{bridgevln}; $^\dagger$ denotes training without navigation data beyond R2R-CE and RxR-CE.
NE is in meters; other scores are percentages.
PanoVLN achieves the highest SR and SPL on both benchmarks.}
    \label{tab:simulation_main}

    \begin{adjustbox}{max width=\tableresize\linewidth}
    \begin{tabular}{@{}l|cccc|cccc|cccc@{}}
        \toprule
        \multirow{2}{*}{Method}
        & \multicolumn{4}{c|}{Observation}
        & \multicolumn{4}{c|}{R2R Val-Unseen}
        & \multicolumn{4}{c}{RxR Val-Unseen} \\
        \cmidrule(lr){2-5}
        \cmidrule(lr){6-9}
        \cmidrule(lr){10-13}
        & Pano. & Odo. & Depth & S.RGB
        & NE$\downarrow$ & OS$\uparrow$ & SR$\uparrow$ & SPL$\uparrow$
        & NE$\downarrow$ & SR$\uparrow$ & SPL$\uparrow$ & nDTW$\uparrow$ \\
        \midrule

        CMA$^*$~\cite{bridgevln}
        & \checkmark & \checkmark & \checkmark &
        & 6.20 & 52.0 & 41.0 & 36.0
        & 8.76 & 26.5 & 22.1 & 47.0 \\

        GridMM$^*$~\cite{gridmm}
        & \checkmark & \checkmark & \checkmark &
        & 5.11 & 61.0 & 49.0 & 41.0
        & -- & -- & -- & -- \\

        ETPNav$^*$~\cite{etpnav}
        & \checkmark & \checkmark & \checkmark &
        & 4.71 & 65.0 & 57.0 & 49.0
        & 5.64 & 54.7 & 44.8 & 61.9 \\

        HNR$^*$~\cite{hnr}
        & \checkmark & \checkmark & \checkmark &
        & 4.42 & 67.0 & 61.0 & 51.0
        & 5.50 & 56.3 & 46.7 & 63.5 \\

        ScaleVLN$^*$~\cite{scalevln}
        & \checkmark & \checkmark & \checkmark &
        & 4.80 & -- & 55.0 & 51.0
        & -- & -- & -- & -- \\

        \midrule

        InstructNav~\cite{instructnav}
        & \checkmark & \checkmark & \checkmark &
        & 6.89 & -- & 31.0 & 24.0
        & -- & -- & -- & -- \\

        LAW~\cite{law}
        & & \checkmark & \checkmark & \checkmark
        & 6.83 & 44.0 & 35.0 & 31.0
        & 10.90 & 8.0 & 8.0 & 38.0 \\

        CM$^2$~\cite{cm2}
        & & \checkmark & \checkmark & \checkmark
        & 7.02 & 41.5 & 34.3 & 27.6
        & -- & -- & -- & -- \\

        WS-MGMap~\cite{wsmgmap}
        & & \checkmark & \checkmark & \checkmark
        & 6.28 & 47.6 & 38.9 & 34.3
        & -- & -- & -- & -- \\

        CMA~\cite{r2rce}
        & & & \checkmark & \checkmark
        & 7.37 & 40.0 & 32.0 & 30.0
        & -- & -- & -- & -- \\

        \midrule

        MapNav$^\dagger$~\cite{mapnav}
        & & \checkmark & \checkmark & \checkmark
        & 4.93 & 53.0 & 39.7 & 37.2
        & 7.62 & 32.6 & 27.7 & 43.5 \\

        StreamVLN$^\dagger$~\cite{streamvln}
        & & & & \checkmark
        & 5.43 & 62.5 & 52.8 & 47.2
        & 6.72 & 48.6 & 42.5 & 60.2 \\

        Aux-Think$^\dagger$~\cite{auxthink}
        & & & & \checkmark
        & 6.08 & 60.0 & 54.8 & 46.9
        & 6.24 & 52.2 & 40.2 & -- \\

        JanusVLN$^\dagger$~\cite{janusvln}
        & & & & \checkmark
        & 5.17 & 58.0 & 52.8 & 49.2
        & 6.46 & 51.4 & 44.3 & 59.1 \\

        CorrectNav$^\dagger$~\cite{correctnav}
        & & & & \checkmark
        & 4.24 & 67.5 & 65.1 & 62.3
        & 4.09 & 69.3 & 63.3 & \textbf{75.2} \\

        \rowcolor{gray!12}
        \textbf{PanoVLN$^\dagger$~(Ours)}
        & \checkmark & & &
        & 3.10 & 79.7 & 73.9 & 67.9
        & 3.14 & 74.1 & 63.9 & 71.4 \\

        \midrule

        NaVid~\cite{navid}
        & & & & \checkmark
        & 5.47 & 49.1 & 37.4 & 35.9
        & -- & -- & -- & -- \\

        Uni-NaVid~\cite{uninavid}
        & & & & \checkmark
        & 5.58 & 53.3 & 47.0 & 42.7
        & 6.24 & 48.7 & 40.9 & -- \\

        NaVILA~\cite{navila}
        & & & & \checkmark
        & 5.22 & 62.5 & 54.0 & 49.0
        & 6.77 & 49.3 & 44.0 & 58.8 \\

        StreamVLN~\cite{streamvln}
        & & & & \checkmark
        & 4.90 & 63.6 & 56.4 & 50.2
        & 5.65 & 54.4 & 45.4 & 63.7 \\

        JanusVLN~\cite{janusvln}
        & & & & \checkmark
        & 4.78 & 65.2 & 60.5 & 56.8
        & 6.06 & 56.2 & 47.5 & 62.1 \\

        DualVLN~\cite{dualvln}
        & & & & \checkmark
        & 4.05 & 70.7 & 64.3 & 58.5
        & 4.58 & 61.4 & 51.8 & 70.0 \\

        AwareVLN~\cite{awarevln}
        & & & & \checkmark
        & 4.02 & 73.5 & 65.4 & 55.1
        & 3.95 & 67.6 & 56.1 & 65.7 \\

        \rowcolor{gray!12}
        \textbf{PanoVLN~(Ours)}
        & \checkmark & & &
        & \textbf{2.83} & \textbf{83.1} & \textbf{77.3} & \textbf{70.6}
        & \textbf{2.85} & \textbf{78.0} & \textbf{65.9} & 73.3 \\

        \bottomrule
    \end{tabular}
    \end{adjustbox}
\end{table}

\section{EXPERIMENTS}
\label{sec:experiments}

\subsection{Experimental Setup}
\label{sec:experimental_setup}

We evaluate on the R2R-CE and RxR-CE Val-Unseen splits~\citep{r2rce,rxr} in Matterport3D scenes~\citep{mp3d} using Habitat~\citep{habitat}.
We report navigation error (NE), oracle success rate (OS), success rate (SR), success weighted by path length (SPL), and normalized dynamic time warping (nDTW).
OS records whether a trajectory comes within $3\,\mathrm{m}$ of the goal; SR also requires stopping there.
SPL measures path efficiency and nDTW reference-route agreement.
NE is in meters; other scores use a 0--100 scale.

PanoVLN uses RGB-only observations.
The default model combines Qwen3.5-4B~\citep{qwen35vl} with a frozen PanoVGGT encoder~\citep{panovggt} and uses an $H=18$ prediction horizon with CGE for execution.
PanoVLN$^\dagger$ trains on R2R-CE and RxR-CE navigation data; the full model additionally uses our constructed dataset.
Ablation configurations are specified with each study.
Training settings and the navigation prompt are in Appendix~\ref{app:implementation}.

\subsection{Simulation Experiments}
\label{sec:simulation_experiments}

\paragraph{Comparison with prior methods.}
PanoVLN achieves state-of-the-art SR of 77.3\% on R2R-CE and 78.0\% on RxR-CE, exceeding the previous best results by 11.9 and 8.7 percentage points, respectively (Table~\ref{tab:simulation_main}).
PanoVLN$^\dagger$ also leads the restricted-data group in SR and SPL on both benchmarks.
Adding our decision-centric trajectories further improves performance in unseen scenes.

\begin{figure}[t]
    \centering
    \includegraphics[width=0.9\linewidth]{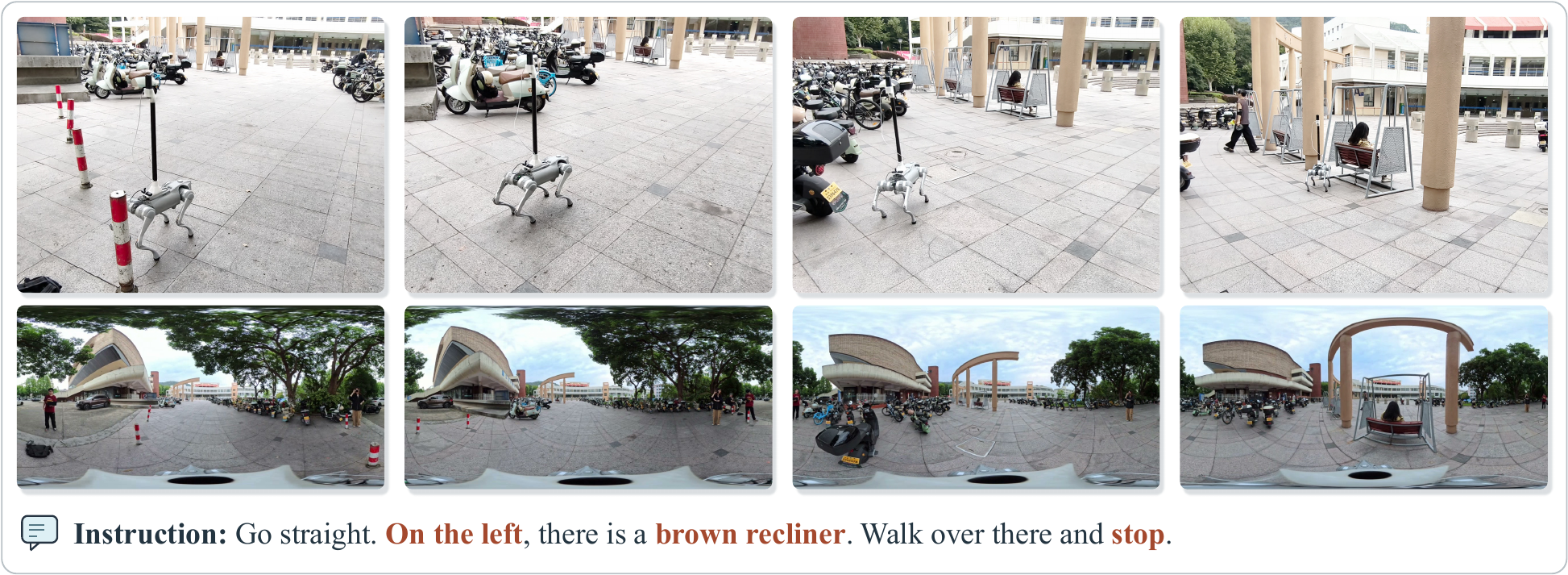}
    \caption{\textbf{Real-world navigation.}
PanoVLN deployed on a quadruped robot and transfers to real-world instruction following without scene-specific fine-tuning.}
    \label{fig:realworld_navigation_cases}
\end{figure}

\paragraph{Use of panoramic context.}
With the same ERP and instruction, short-horizon training concentrates attention ahead, resembling a perspective policy, while PanoVLN attends to instructed passages across directions.
Short targets often share initial movements across routes; longer targets include route choices and subsequent movement, making the distinguishing visual cues relevant to prediction and encouraging use of the full panorama.

\subsection{Real-World Experiments}
\label{sec:realworld_experiments}

\paragraph{Deployment and evaluation.}
All methods use a Unitree Go2, the same Insta360 X5 mounted $1.5\,\mathrm{m}$ above ground, and a remote RTX 3090.
PanoVLN uses approximately $12\,\mathrm{GB}$ of GPU memory; network overhead averages $272\,\mathrm{ms}$ per call, excluding inference.
Execution is synchronous: the robot waits for a response, executes its actions, then requests the next prediction.

We compare NaVid~\citep{navid}, NaVILA~\citep{navila}, StreamVLN~\citep{streamvln}, JanusVLN~\citep{janusvln}, and PanoVLN on 20 shared instruction--route pairs per setting, without scene-specific fine-tuning.
Hallway tests successive turns; Office adds clutter and room transitions.
Campus covers gardens, sports fields, and plazas, testing transfer from indoor training to outdoor spaces and varied terrain.

\begin{figure}[t]
    \centering
    \includegraphics[width=\linewidth]{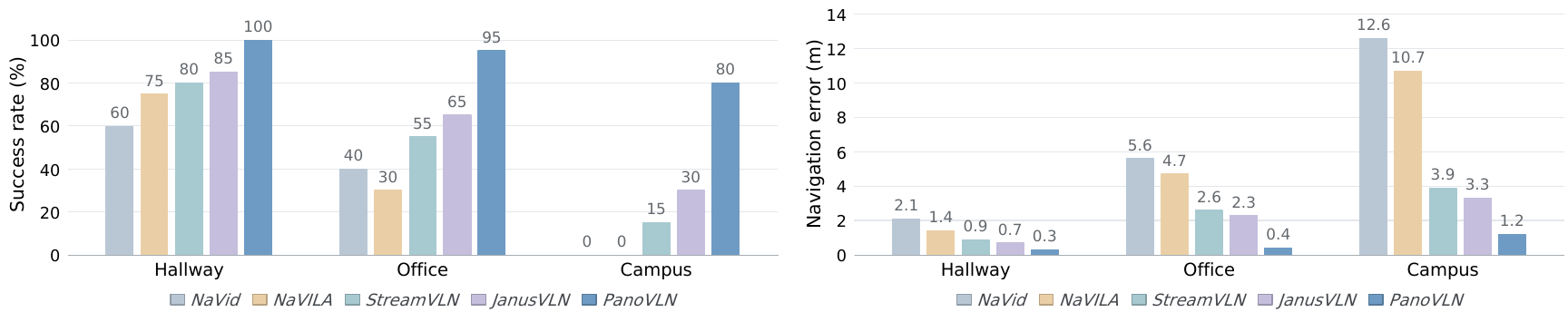}
    \caption{\textbf{Real-world navigation performance.}
We compare SR and NE on 20 shared instruction--route pairs each in Hallway, Office, and Campus and achieves the best navigation performance.}
    \label{fig:realworld_navigation}
\end{figure}

\paragraph{Navigation performance.}
The results test both local route complexity and transfer across scene types (Figure~\ref{fig:realworld_navigation}; qualitative trials are in the supplementary video).
NaVid struggles with instruction progress over long routes, while clutter and room transitions disrupt NaVILA in Office.
StreamVLN and JanusVLN handle indoor routes more reliably but struggle to transfer to open outdoor layouts in Campus.
PanoVLN follows successive indoor route choices and transfers this ability outdoors, maintaining instruction-to-route grounding across changes in appearance and spatial layout.

\begin{table}[t]
    \centering
    \small
    \renewcommand{\arraystretch}{1.0}
    \setlength{\tabcolsep}{4pt}
    \caption{\textbf{Real-world execution efficiency.}
PanoVLN achieves the best overall navigation efficiency among the compared methods.}
    \label{tab:realworld_efficiency}
\begin{tabular}{@{}l@{\hspace{8pt}}cccccc@{}}
            \toprule
            \multirow[t]{2}{*}{Method}
            & \multicolumn{2}{c}{Navigation}
            & \multicolumn{2}{c}{Continuity}
            & \multicolumn{2}{c}{Planning} \\
            \cmidrule(lr){2-3}
            \cmidrule(lr){4-5}
            \cmidrule(lr){6-7}
            & Time~(s) $\downarrow$
            & Speed~(cm/s) $\uparrow$
            & Wait~(\%) $\downarrow$
            & Pauses $\downarrow$
            & Calls $\downarrow$
            & Latency~(s) $\downarrow$ \\
            \midrule
            NaVid
            & 135.2 & 12.9 & 23.2 & 15.7 & 29.4 & 0.90 \\
            NaVILA
            & 194.0 & 8.2 & 24.7 & 29.6 & 41.3 & 1.05 \\
            StreamVLN
            & 117.8 & 13.3 & 27.1 & 8.3 & 34.3 & \textbf{0.58} \\
            JanusVLN
            & 412.9 & 5.3 & 39.2 & 95.3 & 101.7 & 1.32 \\
            \rowcolor{gray!12}
            \textbf{PanoVLN~(Ours)}
            & \textbf{86.4} & \textbf{25.7} & \textbf{13.6}
            & \textbf{5.4} & \textbf{7.4} & 1.08 \\
            \bottomrule
        \end{tabular}
\end{table}

\paragraph{Execution efficiency.}
Table~\ref{tab:realworld_efficiency} reports trial averages (see Appendix~\ref{app:execution_metrics}).
\emph{Time} is navigation duration; \emph{Speed} is traveled distance divided by duration, including waiting.
\emph{Wait} is the fraction of time awaiting policy responses; \emph{Pauses} counts stationary intervals longer than $1\,\mathrm{s}$.
\emph{Calls} counts policy requests; \emph{Latency} is inference time per request, excluding network communication.

Under synchronous execution, frequent policy requests add inference and communication delay and interrupt motion.
JanusVLN requests a prediction for each action; StreamVLN has lower per-call latency but replans every four actions.
PanoVLN predicts longer segments, and CGE selects a confident prefix before requesting a new observation.
Sustaining motion while predictions remain confident reduces interruptions and total navigation time.

\setcounter{topnumber}{3}

\begin{figure}[t]
    \centering
    \includegraphics[width=0.9\linewidth]{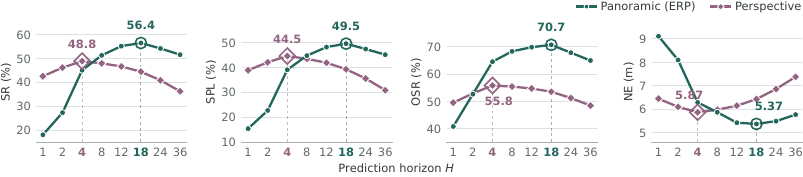}
    \caption{\textbf{Prediction horizon ablation.}
Panoramic policies benefit from longer supervision and outperform perspective policies at longer horizons.}
    \label{fig:action_horizon}
\end{figure}

\subsection{Ablation Studies}
\label{sec:ablation_studies}

The horizon, execution, sampling, and geometry studies use R2R-CE and RxR-CE training data; the data study varies the additional trajectories.
Evaluation uses the corresponding Val-Unseen splits.

\paragraph{Prediction horizon.}
We vary the prediction horizon while fixing the VLM, visual-token budget, and random-start sampling, training geometry-free policies for 4,000 updates (Figure~\ref{fig:action_horizon}).
ERP policies trail perspective policies at short horizons but overtake them as targets lengthen.
Perspective performance peaks at $H=4$, while ERP favors $H=18$.

Short targets may end before visible routes diverge; longer ERP targets supervise the intended choice.
Longer perspective targets increasingly require cues outside the current view.
ERP improves from $H=8$ to $H=18$ with execution fixed at six actions, linking the gain to longer supervision.
These results favor matching supervision to visible route information; we use $H=18$ thereafter.

\noindent\begin{minipage}{\linewidth}
\begin{minipage}[t]{0.57\linewidth}
\vspace{0pt}
\textbf{Training-state sampling.}
We compare random starts with our turn- and termination-aware sampling.
Both variants use a geometry-free $H=18$ architecture, 4,000 updates, and six-action execution.
The random variant is the $H=18$ ERP run in Figure~\ref{fig:action_horizon}.
\end{minipage}\hfill
\begin{minipage}[t]{0.40\linewidth}
    \vspace{0pt}
    \centering
    \small
    \setlength{\abovecaptionskip}{0pt}
    \setlength{\belowcaptionskip}{5pt}
    \captionof{table}{\textbf{Training data comparison.}}
    \label{tab:sampling_ablation}
    \setlength{\tabcolsep}{3pt}
    \renewcommand{\arraystretch}{1.1}
\begin{tabular}{@{}lrrrr@{}}
        \toprule
        Sampling & NE$\downarrow$ & OS$\uparrow$ & SR$\uparrow$ & SPL$\uparrow$ \\
        \midrule
        Random & 5.37 & \textbf{70.7} & 56.4 & 49.5 \\
        \rowcolor{gray!10}
        Ours & \textbf{4.40} & 69.9 & \textbf{62.8} & \textbf{57.2} \\
        \bottomrule
    \end{tabular}
\end{minipage}
\end{minipage}\par

Long stretches of forward motion supply similar targets, while brief turn and stop states determine route transitions and completion.
Our sampling gives these states more supervision (Table~\ref{tab:sampling_ablation}).
Comparable OS and higher SR indicate more reliable termination in the goal region, highlighting the value of learning when to change or end an action sequence.

\begin{figure}[t]
    \centering
    \includegraphics[width=\linewidth]{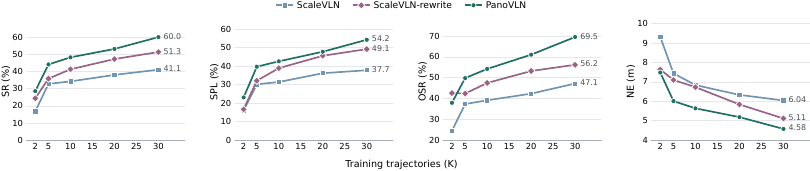}
    \caption{\textbf{Training data comparison.}
We compare data sources across training scales on R2R-CE Val-Unseen. Our data consistently delivers the best navigation performance.}
    \label{fig:data_scale_ablation}
\end{figure}
\begin{table}[t]
    \centering
    \small
    \renewcommand{\arraystretch}{1.0}
    \caption{\textbf{Execution strategy comparison.}
We compare fixed, random, and confidence-guided execution.
CGE achieves the best navigation performance on both benchmarks.}
    \label{tab:cge_ablation}
    \begin{adjustbox}{max width=\tableresize\linewidth}
    \begin{tabular}{@{}l*{8}{c}@{}}
        \toprule
        \multirow{2}{*}{Execution strategy}
        & \multicolumn{4}{c}{R2R-CE}
        & \multicolumn{4}{c}{RxR-CE} \\
        \cmidrule(lr){2-5}\cmidrule(lr){6-9}
        & NE$\downarrow$ & OS$\uparrow$ & SR$\uparrow$ & SPL$\uparrow$
        & NE$\downarrow$ & SR$\uparrow$ & SPL$\uparrow$ & nDTW$\uparrow$ \\
        \midrule
        Fixed: 1 action & 4.42 & 68.2 & 64.6 & 60.5 & 4.11 & 65.7 & 55.6 & 70.3 \\
        Fixed: 6 actions & 4.43 & 70.8 & 64.2 & 58.7 & 4.25 & 65.4 & 54.2 & 69.0 \\
        Fixed: 12 actions & 4.65 & 69.7 & 61.4 & 55.4 & 4.98 & 61.1 & 46.4 & 64.6 \\
        Fixed: 18 actions & 4.89 & 67.0 & 57.2 & 50.9 & 5.64 & 54.9 & 43.4 & 59.8 \\
        Random: 1--18 actions & 4.62 & 68.5 & 61.3 & 55.6 & 5.07 & 59.9 & 52.3 & 64.9 \\
        \rowcolor{gray!12}
        \textbf{CGE~(Ours)} & \textbf{4.10} & \textbf{71.0} & \textbf{66.6} & \textbf{61.5} & \textbf{4.06} & \textbf{66.5} & \textbf{56.4} & \textbf{70.8} \\
        \bottomrule
    \end{tabular}
    \end{adjustbox}
\end{table}

\paragraph{Confidence-guided execution.}
Table~\ref{tab:cge_ablation} compares fixed, random, and confidence-guided execution using the same geometry-free $H=18$ checkpoint, trained for one epoch with our sampling.

Long fixed prefixes commit to increasingly uncertain actions, while random prefixes ignore the model's confidence.
CGE adapts execution to each state, outperforming random prefixes and all fixed strategies, including one-action execution.
These studies separate two choices: long targets teach route selection, while CGE determines when to reobserve.

\begin{table}[t]
    \centering
    \small
    \renewcommand{\arraystretch}{1.05}
    \caption{\textbf{Geometry encoder comparison.}
We compare geometric features from different encoders.
PanoVGGT yields the highest success rates on both benchmarks.}
    \label{tab:spatial_encoder_ablation}
    \begin{adjustbox}{max width=\tableresize\linewidth}
    \begin{tabular}{@{}l*{8}{c}@{}}
        \toprule
        \multirow{2}{*}{Extra encoder}
        & \multicolumn{4}{c}{R2R-CE}
        & \multicolumn{4}{c}{RxR-CE} \\
        \cmidrule(lr){2-5}\cmidrule(lr){6-9}
        & NE$\downarrow$ & OS$\uparrow$ & SR$\uparrow$ & SPL$\uparrow$
        & NE$\downarrow$ & SR$\uparrow$ & SPL$\uparrow$ & nDTW$\uparrow$ \\
        \midrule
        None
        & 4.10 & 71.0 & 66.6 & 61.5 & 4.06 & 66.5 & 56.4 & 70.8 \\
        UniK3D
        & \textbf{3.77} & 73.2 & 67.8 & 63.3 & \textbf{3.92} & 65.4 & \textbf{57.3} & 68.1 \\
        DA$^2$
        & 4.01 & 72.3 & 65.5 & 61.6 & 4.35 & 63.0 & 55.9 & 67.7 \\
        DAP
        & 4.15 & 70.9 & 64.5 & 60.4 & 4.48 & 61.1 & 54.3 & 66.5 \\
        \rowcolor{gray!12}
        \textbf{PanoVGGT~(Ours)}
        & 3.91 & \textbf{73.6} & \textbf{68.6} & \textbf{63.7}
        & 4.02 & \textbf{67.1} & 57.2 & \textbf{71.5} \\
        \bottomrule
    \end{tabular}
    \end{adjustbox}
\end{table}

\paragraph{Geometric features.}
We compare frozen UniK3D~\citep{unik3d}, DA$^2$~\citep{da2}, DAP~\citep{dap}, and PanoVGGT encoders with the same fusion design, visual-token budget, and CGE (Table~\ref{tab:spatial_encoder_ablation}).
The alternative encoders have mixed effects, whereas PanoVGGT improves SR on both benchmarks.
Its panoramic features link landmarks to neighboring passages and directions, complementing semantic appearance with spatial cues for route selection.

\newpage
\paragraph{Training data.}
We compare ScaleVLN~\citep{scalevln}, ScaleVLN-rewrite, and our data at matched trajectory counts, using PanoVGGT and CGE without DAgger~\citep{dagger} (Figure~\ref{fig:data_scale_ablation}).
ScaleVLN-rewrite retains the original routes and regenerates instructions using our instruction pipeline.
Its gains over ScaleVLN demonstrate the benefit of improved instruction quality.
Our full pipeline yields further gains across training scales by combining clear instructions with decision-rich routes, providing repeated supervision for language-guided selection among paths visible in a panorama.
Appendices~\ref{app:data_composition} and~\ref{app:backbone_comparison} provide data-composition and backbone comparisons.

\section{Conclusion}
We presented PanoVLN for vision-and-language navigation from RGB panoramas, combining longer action supervision, confidence-guided execution, decision-centric training, and geometric features.
With a 4B backbone, it achieves state-of-the-art success rates on R2R-CE and RxR-CE and enables faster real-world navigation with fewer pauses.
Our findings suggest that broader visibility should shape how navigation policies learn and act.

\clearpage
\begingroup
\interlinepenalty=10000 
\bibliography{ref}
\bibliographystyle{iclr2027_conference}
\endgroup

\appendix
\clearpage
\section{THE USE OF LARGE LANGUAGE MODELS}
\label{app:llm_usage}

We used GPT-5.6 and GPT-6 to edit manuscript language and layout.
The authors review all edits and are responsible for the technical content, experimental results, and conclusions.

For dataset construction, Qwen3.8-27B~\citep{qwen38} generates instructions from trajectory videos and compass images.
We verify these against the observations, revise incorrect route or stopping descriptions, and discard samples that fail verification~(Section~\ref{sec:training_data}).

\section{IMPLEMENTATION DETAILS}
\label{app:implementation}
\label{app:training_config}

\subsection{Model and training settings}
\label{app:model_training_settings}

Table~\ref{tab:training_configuration} lists model and training settings; Figure~\ref{fig:inference_prompt} gives the training and inference prompt.

\begin{table}[!ht]
\centering
\footnotesize
\renewcommand{\arraystretch}{1.04}
\setlength{\tabcolsep}{5pt}
\caption{Default PanoVLN settings.
    Image dimensions are width $\times$ height.}
\label{tab:training_configuration}
\begin{tabularx}{\linewidth}{@{}>{\raggedright\arraybackslash}p{0.42\linewidth}>{\raggedright\arraybackslash}X@{}}
\toprule
Setting & Value \\
\midrule
VLM / frozen geometry encoder & Qwen3.5-4B / PanoVGGT \\
Geometry features / fusion & Last aggregator layer; $2\times2$ grouping; fusion after the visual merger \\
Geometry normalization / projection & RMSNorm / MLP~($8192\rightarrow4096\rightarrow2560$, GELU) \\
Residual scale & $\alpha=0.2$ \\
Current / history ERP resize & $960\times480$ / $448\times224$, before cropping \\
Visual-encoder ERP crop & $20^\circ$ from each pole \\
Geometry-encoder input & $1036\times518$; full panorama \\
Visual history & Up to 10 earlier frames from the latest 100 observations \\
\midrule
Optimizer / epochs & Fused AdamW / 1 \\
Effective batch size & 128~(8 GPUs $\times$ 4 examples $\times$ 4 accumulation steps) \\
Vision-encoder learning rate & $2\times10^{-6}$ \\
Language / merger / projector LR & $2\times10^{-5}$ \\
Learning-rate schedule / warmup & Cosine / $3\%$ of training steps \\
Weight decay / gradient clipping & $0.01$ / max norm $10$ \\
Precision / random seed & bfloat16 / 42 \\
Gradient checkpointing & Enabled \\
\midrule
Prediction horizon / decoding & 18 actions / greedy \\
CGE budget / minimum execution & $B=1.2$ / $E_{\min}=4$ actions \\
Action space & \texttt{forward}: 25 cm; \texttt{left}/\texttt{right}: $15^\circ$; \texttt{stop} \\
\bottomrule
\end{tabularx}
\end{table}

\begin{figure}[!ht]
\centering
\setlength{\fboxsep}{7pt}
\fbox{\parbox{\dimexpr\linewidth-2\fboxsep-2\fboxrule\relax}{
\footnotesize
\textbf{System}\par\smallskip
You are an autonomous navigation assistant.
Your task is to follow the navigation instruction.
Given the instruction, your recent observations, and your current observation, devise an action sequence using the four actions: left or right by 15 degrees, forward by 25 centimeters, or stop once the task is complete.
Return exactly 18 action words in execution order, separated by spaces.
If the task is complete before 18 actions, fill the remaining positions with stop.

\medskip
\textbf{User}\par\smallskip
Instruction: \emph{[navigation instruction]}\\
History memory observations are panoramic views ordered from older to newer:\\
\emph{[history panorama 1]} $\cdots$ \emph{[history panorama $m$]}\\
Current observation~(panoramic view):\\
\emph{[current panorama]}\\
Devise the next action sequence.
}}
\caption{Navigation prompt~($H=18$).
    Italic fields are inputs; history is omitted when unavailable.}
\label{fig:inference_prompt}
\end{figure}

\clearpage
\subsection{Real-world execution metrics}
\label{app:execution_metrics}

Table~\ref{tab:realworld_efficiency} summarizes the time and planning costs of complete navigation trials under synchronous execution (Sec.~\ref{sec:realworld_experiments}).
The robot requests a prediction, waits for the response, executes the selected actions, and then starts the next request.
We compute the six metrics for each trial before averaging them across trials.

\paragraph{Navigation duration and speed.}
For trial $i$, let $T_i$ be its elapsed navigation time in seconds and $L_i$ its traveled distance in meters.
The two navigation metrics are
\begin{equation}
    \mathrm{Time}_i = T_i,
    \qquad
    \mathrm{Speed}_i = \frac{100L_i}{T_i}\;\mathrm{cm/s}.
\end{equation}
Time includes both action execution and waiting for policy responses.
Speed measures progress along the executed trajectory per unit of elapsed time, with waiting included in the denominator.
It therefore captures the combined effect of robot motion and interruptions during the trial.

\paragraph{Waiting time and motion interruptions.}
Let $K_i$ be the number of policy requests and $w_{ij}$ the elapsed time from issuing request $j$ to receiving its response.
Total policy-waiting time is $W_i=\sum_{j=1}^{K_i}w_{ij}$, giving
\begin{equation}
    \mathrm{Wait}_i = 100\frac{W_i}{T_i}\;\%.
\end{equation}
Each waiting interval includes model inference and network communication.
Wait measures the fraction of the trial consumed by these response intervals.
Pauses counts continuous stationary intervals lasting longer than $1\,\mathrm{s}$ during the trial.
Writing their durations as $d_{ik}$,
\begin{equation}
    \mathrm{Pauses}_i = \sum_k \mathbf{1}[d_{ik}>1\,\mathrm{s}].
\end{equation}
A continuous interval is counted once.
Wait describes the duration of policy-related delays, whereas Pauses describes the frequency of sustained motion interruptions.
A response received within one second contributes to Wait even when it creates no pause longer than the threshold.

\paragraph{Policy calls and inference latency.}
Calls counts the requests issued during a trial.
Let $\ell_{ij}$ be the model inference time for request $j$ on the RTX 3090.
We compute
\begin{equation}
    \mathrm{Calls}_i = K_i,
    \qquad
    \mathrm{Latency}_i = \frac{1}{K_i}\sum_{j=1}^{K_i}\ell_{ij}\;\mathrm{s}.
\end{equation}
Latency measures server-side model inference; the request--response interval used for Wait also includes communication.
The measured mean network overhead is $272\,\mathrm{ms}$ per call.
Calls and Latency describe the frequency and duration of planning, respectively, and together explain the policy-related waiting accumulated during navigation.

\paragraph{Aggregation.}
For each method, the table reports the arithmetic mean of each trial-level metric over the evaluated trials.
Speed and Wait are computed separately for each trial and then averaged.
Latency is first averaged over requests within a trial, then over trials.
Pauses and Calls are integer counts for individual trials; their reported means can be fractional.

\clearpage
\section{MORE ABLATION STUDIES}
\label{app:more_ablations}

These studies complement Sec.~\ref{sec:ablation_studies} by examining training-source composition and the VLM backbone.
Both use the R2R-CE and RxR-CE Val-Unseen evaluation protocol in Sec.~\ref{sec:experimental_setup}.

\subsection{Training data composition}
\label{app:data_composition}

Table~\ref{tab:data_ablation} adds training sources while holding Qwen3.5-4B, PanoVGGT, CGE, $H=18$, and sample selection fixed.
Starting from R2R-CE and RxR-CE, DAgger raises SR by 5.3 and 7.0 points and SPL by 4.2 and 6.7 points, respectively.
This configuration corresponds to PanoVLN$^\dagger$ in Table~\ref{tab:simulation_main}.
Adding our constructed dataset further improves SR by 3.4/3.9 points and SPL by 2.7/2.0 points, yielding the full PanoVLN model.
RxR-CE nDTW also increases from 71.4 to 73.3, indicating closer agreement with the instructed route alongside better completion.

The two added sources therefore provide complementary gains: our decision-centric data remains useful after DAgger has already strengthened the policy.
This cumulative comparison assesses the full training recipe; Figure~\ref{fig:data_scale_ablation} complements it by comparing data construction strategies at matched trajectory counts.

\begin{table}[!ht]
    \centering
    \small
    \renewcommand{\arraystretch}{1.2}
    \setlength{\tabcolsep}{3pt}
    \caption{Training data composition on R2R-CE and RxR-CE Val-Unseen.
        All settings retain PanoVGGT, CGE, and the $H=18$ prediction setting.}
    \label{tab:data_ablation}
    \begin{adjustbox}{max width=\tableresize\linewidth}
    \begin{tabular}{@{}*{12}{c}@{}}
        \toprule
        \multicolumn{4}{c}{Training data}
        & \multicolumn{4}{c}{R2R-CE}
        & \multicolumn{4}{c}{RxR-CE} \\
        \cmidrule(lr){1-4}
        \cmidrule(lr){5-8}
        \cmidrule(lr){9-12}

        R2R & RxR & DAgger & PanoVLN
        & NE$\downarrow$ & OS$\uparrow$ & SR$\uparrow$ & SPL$\uparrow$
        & NE$\downarrow$ & SR$\uparrow$ & SPL$\uparrow$ & nDTW$\uparrow$ \\
        \midrule

        $\checkmark$ & $\checkmark$ & &
        & 3.91 & 73.6 & 68.6 & 63.7
        & 4.02 & 67.1 & 57.2 & 71.5 \\

        $\checkmark$ & $\checkmark$ & $\checkmark$ &
        & 3.10 & 79.7 & 73.9 & 67.9
        & 3.14 & 74.1 & 63.9 & 71.4 \\

        $\checkmark$ & $\checkmark$ & $\checkmark$ & $\checkmark$
        & 2.83 & 83.1 & 77.3 & 70.6
        & 2.85 & 78.0 & 65.9 & 73.3 \\

        \bottomrule
    \end{tabular}
    \end{adjustbox}
\end{table}

\subsection{Vision-language backbone}
\label{app:backbone_comparison}

Table~\ref{tab:backbone_ablation} evaluates the backbone choice~\citep{qwen3vl,qwen25vl} with all four training sources, PanoVGGT, CGE, visual history, and $H=18$ held fixed.
Qwen3-VL-4B already achieves 76.2\% and 75.7\% SR on R2R-CE and RxR-CE, respectively, with corresponding SPL scores of 70.4\% and 64.9\%.
Replacing it with Qwen3.5-4B improves SR by 1.1 and 2.3 points, respectively, and also improves the remaining reported metrics.
These modest, consistent gains motivate our default backbone, while the strong Qwen3-VL-4B results show that PanoVLN's performance is not confined to a single VLM backbone.

\begin{table}[!ht]
    \centering
    \small
    \renewcommand{\arraystretch}{1.2}
    \setlength{\tabcolsep}{4pt}
    \caption{VLM backbone comparison on R2R-CE and RxR-CE Val-Unseen.
        All backbones use the full training-data configuration, PanoVGGT, CGE, and $H=18$.}
    \label{tab:backbone_ablation}
    \begin{adjustbox}{max width=\tableresize\linewidth}
    \begin{tabular}{@{}l*{8}{c}@{}}
        \toprule
        Backbone
        & \multicolumn{4}{c}{R2R-CE}
        & \multicolumn{4}{c}{RxR-CE} \\
        \cmidrule(lr){2-5}\cmidrule(lr){6-9}
        & NE$\downarrow$ & OS$\uparrow$ & SR$\uparrow$ & SPL$\uparrow$
        & NE$\downarrow$ & SR$\uparrow$ & SPL$\uparrow$ & nDTW$\uparrow$ \\
        \midrule
        Qwen2.5-VL-7B
        & 2.96 & 82.2 & 75.9 & 68.6
        & 2.88 & 76.3 & 65.1 & 72.7 \\

        Qwen3-VL-4B
        & 2.91 & 81.5 & 76.2 & 70.4
        & 2.96 & 75.7 & 64.9 & 71.6 \\

        \rowcolor{gray!10}
        Qwen3.5-4B
        & 2.83 & 83.1 & 77.3 & 70.6
        & 2.85 & 78.0 & 65.9 & 73.3 \\
        \bottomrule
    \end{tabular}
    \end{adjustbox}
\end{table}

\clearpage
\section{TRAINING DATA ANALYSIS}
\label{app:training_data_analysis}
\label{app:dataset_statistics}
\label{app:path_turns}

\begin{figure}[!ht]
    \centering
    \includegraphics[width=0.75\linewidth]{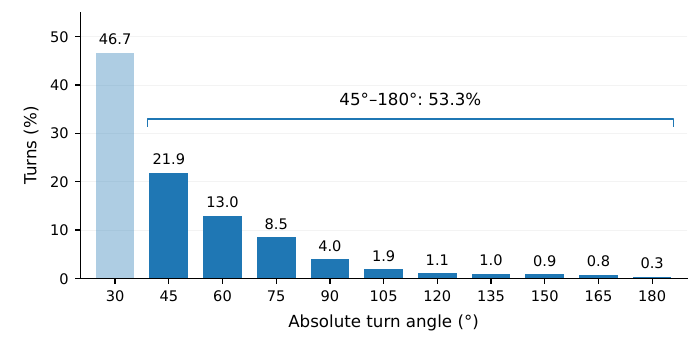}
    \caption{Reference-path turn distribution in R2R-CE and RxR-CE, normalized over 54,457 turns of at least $30^\circ$.
        The $45^\circ$--$180^\circ$ categories account for 53.3\% of turns.}
    \label{fig:reference_path_turns}
\end{figure}

\paragraph{Reference-path turns.}
Figure~\ref{fig:reference_path_turns} summarizes directional changes along reference paths in R2R-CE and RxR-CE.
The pooled distribution uses $15^\circ$ angle categories and is normalized over 54,457 turns of at least $30^\circ$.
The $45^\circ$--$180^\circ$ categories account for 53.3\% of these turns, showing that substantial direction changes are common in the reference routes.

For a forward $90^\circ$ view aligned with the incoming path, an outgoing direction beyond $45^\circ$ lies outside the current field of view.
ERP observations retain these directions before rotation, providing visual context for the route beyond the turn.

\clearpage
\section{QUALITATIVE RESULTS}
\label{app:more_qualitative}

\subsection{Paired panoramic attention cases}
\label{app:attention_protocol}

\begin{figure}[!ht]
    \centering
    \includegraphics[width=0.96\linewidth,height=190mm,keepaspectratio]{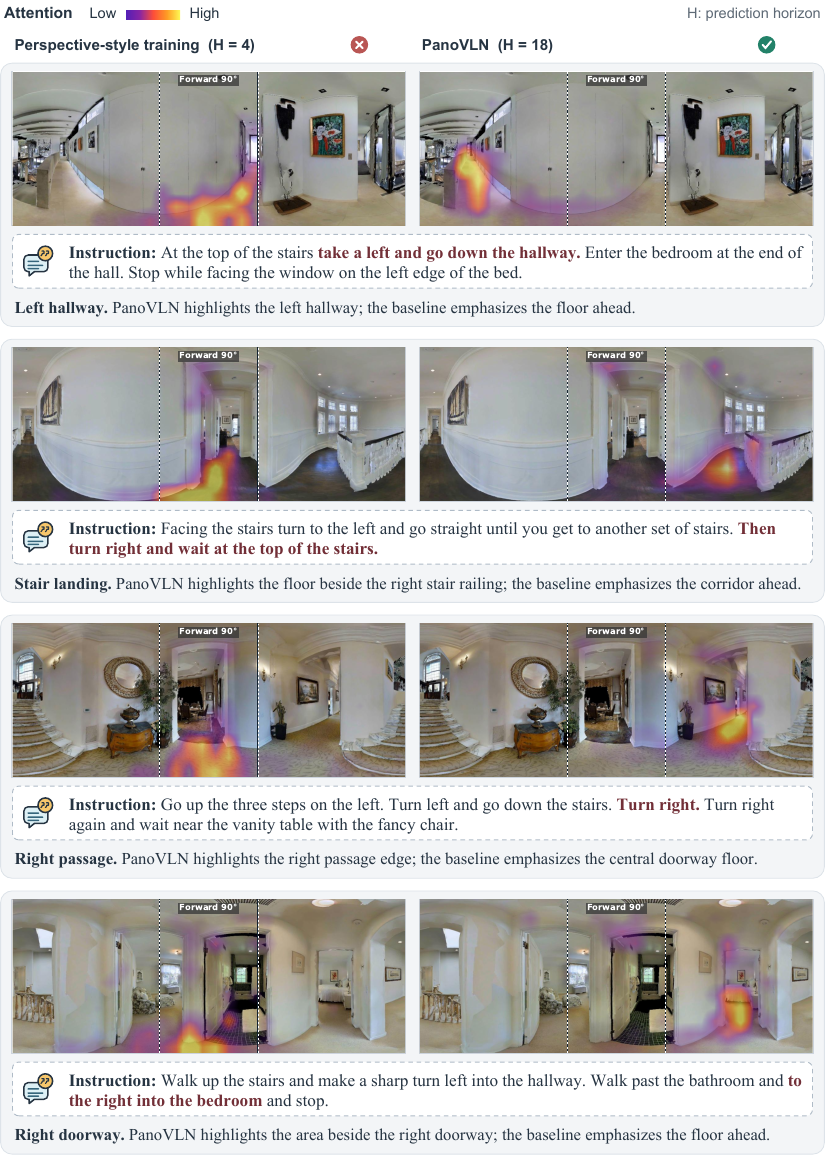}
    \caption{Additional paired attention cases, each using the same ERP and instruction.
        Red bold text marks the relevant route clause; dashed lines bound the forward $90^\circ$ sector.
        Each case note identifies the visible hotspot locations.}
    \label{fig:attention_gallery}
    \label{fig:attention_additional}
    \label{fig:attention_case_hallway}
    \label{fig:attention_case_landing}
    \label{fig:attention_case_passage}
    \label{fig:attention_case_bedroom}
\end{figure}

\clearpage
\subsection{Real-world navigation cases}
\label{app:navigation_cases}
\label{app:robot_examples}

\begin{figure}[!ht]
    \centering
    \panovlnfigure{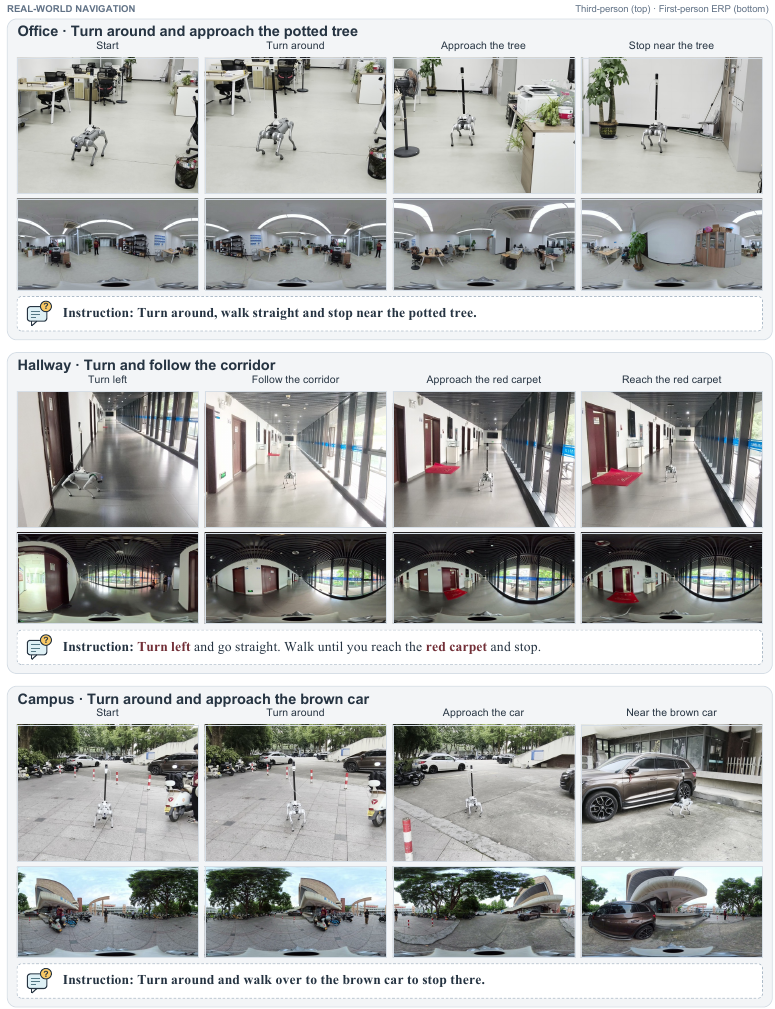}{195mm}
    \caption{Real-world navigation cases in Hallway, Office, and Campus.
        Each row shows the instruction and successive observations from one trial, with route annotations.}
    \label{fig:robot_examples}
    \label{fig:realworld_qualitative}
    \label{fig:realworld_demo}
\end{figure}

\clearpage
\subsection{R2R-CE navigation cases}
\label{app:r2r_examples}

\begin{figure}[!ht]
    \centering
    \panovlnfigure{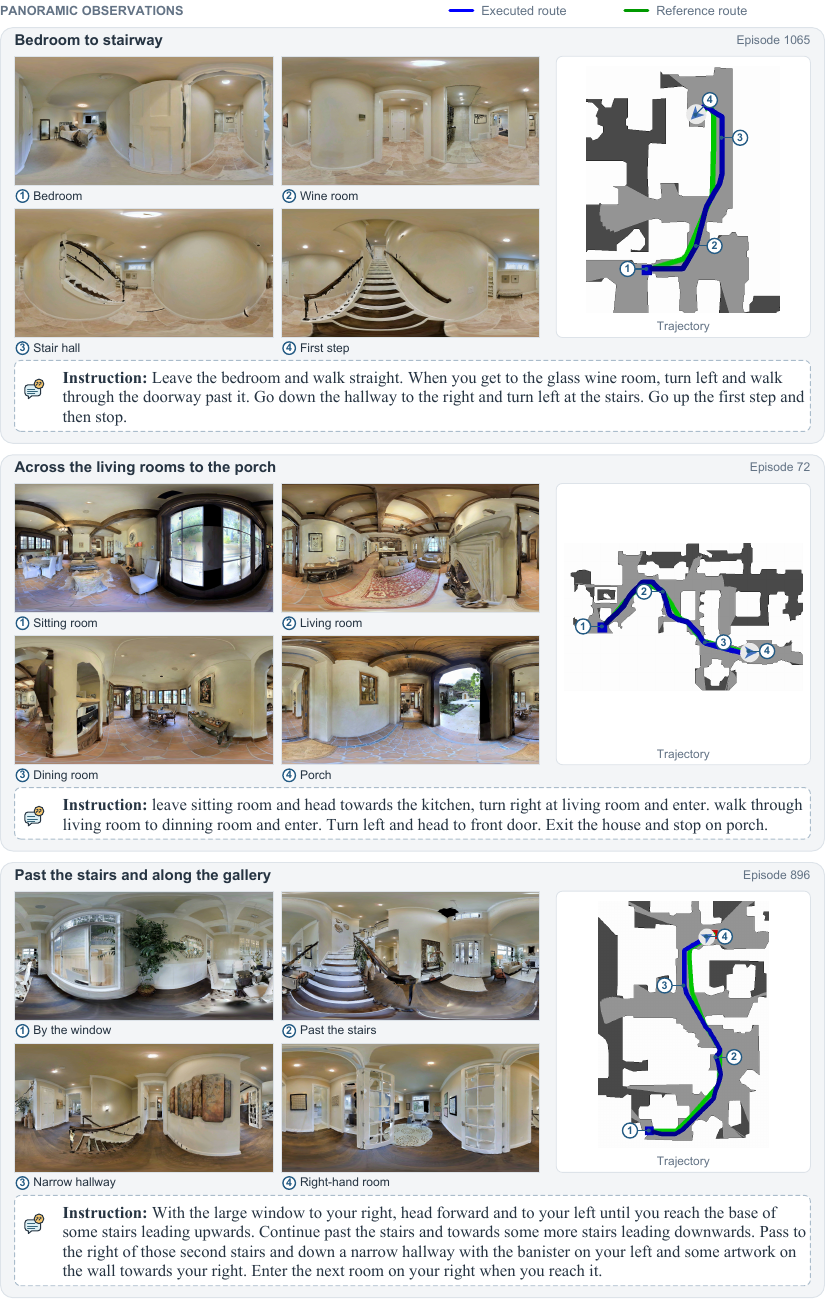}{190mm}
    \caption{Navigation cases on R2R-CE Val-Unseen.
        Each case shows four panoramic observations in reading order and the full instruction.
        Numbered markers link each observation to its location on the trajectory; blue and green denote the executed and reference routes.}
    \label{fig:r2r_examples}
\end{figure}

\clearpage
\subsection{RxR-CE navigation cases}
\label{app:rxr_examples}

\begin{figure}[!ht]
    \centering
    \panovlnfigure{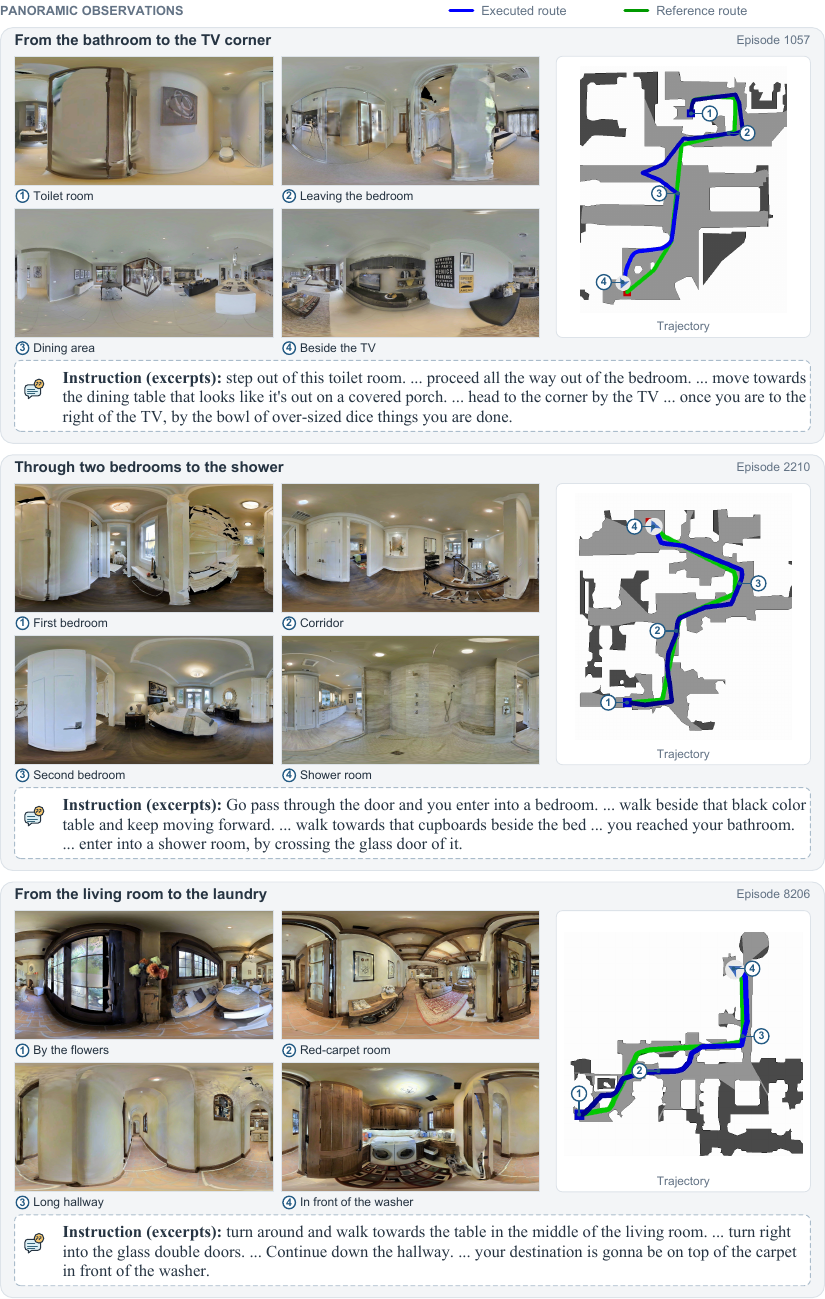}{190mm}
    \caption{Navigation cases on RxR-CE Val-Unseen.
        Four panoramic observations per case are linked to the trajectory by numbered markers.
        Blue and green denote the executed and reference routes.
        Instruction excerpts retain the original wording; ellipses mark omissions.}
    \label{fig:rxr_examples}
\end{figure}

\end{document}